\documentclass[journal]{IEEEtran}
\ifCLASSINFOpdf
  \usepackage[pdftex]{graphicx}
  \graphicspath{{figures/}}
  \DeclareGraphicsExtensions{.pdf,.jpeg,.png,.bmp}
\else
\fi
\usepackage{amsmath}
\usepackage{amssymb}

\usepackage{array}

\ifCLASSOPTIONcompsoc
  \usepackage[caption=false,font=normalsize,labelfont=sf,textfont=sf]{subfig}
\else
  \usepackage[caption=false,font=footnotesize]{subfig}
\fi

\usepackage{multirow}

\usepackage{url}

\usepackage{tikz}
\usetikzlibrary{shapes,arrows,positioning,calc}

\newcommand\copyrighttext{%
  \centering\footnotesize This work has been submitted to the IEEE for possible publication. Copyright may be transferred without notice, after which this version may no longer be accessible.}
\newcommand\copyrightnotice{%
\begin{tikzpicture}[remember picture,overlay]
\node[anchor=south,yshift=-2.2cm] at (current page.north) {\fbox{\parbox{\dimexpr0.9\textwidth-\fboxsep-\fboxrule\relax}{\copyrighttext}}};
\end{tikzpicture}%
}

\begin{document}
%
\title{LiDAR-Based Vehicle Detection and Tracking for Autonomous Racing}
%
%
%

\author{Marcello~Cellina,   
        Matteo~Corno, and~Sergio~Matteo~Savaresi
\thanks{

Marcello Cellina (corresponding author), Matteo Corno, and Sergio Matteo Savaresi are with the Dipartimento di Elettronica, Informazione e Bioingegneria, Politecnico di Milano, Piazza Leonardo da Vinci 32, 20133 Milan, Italy. Email: 
\tt\small{\{marcello.cellina, matteo.corno, sergio.savaresi\}@polimi.it}
}

\thanks{Manuscript received April 19, 2005; revised August 26, 2015.}}

%
%

\markboth{Journal of \LaTeX\ Class Files,~Vol.~14, No.~8, August~2015}%
{Shell \MakeLowercase{\textit{et al.}}: Bare Demo of IEEEtran.cls for IEEE Journals}
%



\maketitle

\copyrightnotice

\begin{abstract}

Autonomous racing provides a controlled environment for testing the software and hardware of autonomous vehicles operating at their performance limits. Competitive interactions between multiple autonomous racecars however introduce challenging and potentially dangerous scenarios.
Accurate and consistent vehicle detection and tracking is crucial for overtaking maneuvers, and low-latency sensor processing is essential to respond quickly to hazardous situations.
This paper presents the LiDAR-based perception algorithms deployed on Team PoliMOVE's autonomous racecar, which won multiple competitions in the Indy Autonomous Challenge series.
Our Vehicle Detection and Tracking pipeline is composed of a novel fast Point Cloud Segmentation technique and a specific Vehicle Pose Estimation methodology, together with a variable-step Multi-Target Tracking algorithm.
Experimental results demonstrate the algorithm's performance, robustness, computational efficiency, and suitability for autonomous racing applications, enabling fully autonomous overtaking maneuvers at velocities exceeding 275 km/h.

\end{abstract}


%
\IEEEpeerreviewmaketitle

\section{Introduction}\label{sec: introduction}

%
%
%



\IEEEPARstart{A}{utonomous racing} allows for safe testing of an autonomous vehicle's full software and hardware stack at the limits of its performance in a controlled environment. 

The competitive interaction of multiple autonomous racecars drastically increases the occurrence of challenging situations like high-speed obstacles and collision avoidance \cite{wischnewski2022indy}. Providing this kind of testing environment is one of the main goals of the Indy Autonomous Challenge (IAC), the first multi-vehicle competition series for level 4 autonomous racecars.

\begin{figure}[h]
    \centering
    \includegraphics[width=\linewidth]{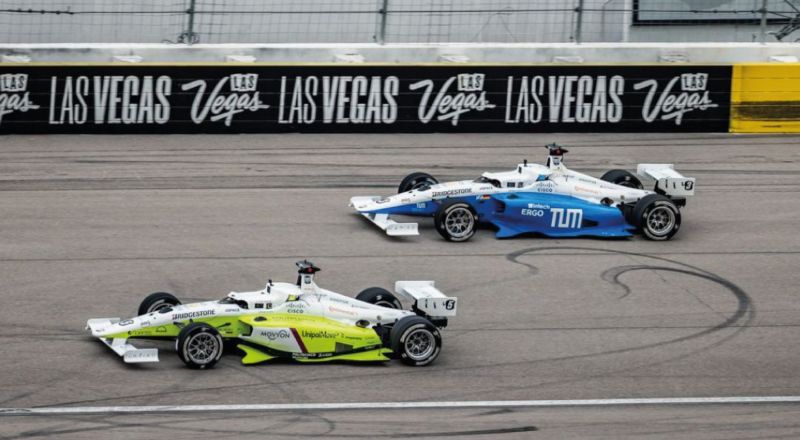}
    \caption{Team PoliMOVE's Dallara AV21 ``MinerVa'' defending from an autonomous overtaking maneuver initiated by TUM Autonomous Motorsport during the final race of the Indy Autonomous Challenge event on January 7, 2023, at the Las Vegas Motor Speedway \textit{Credits: Indy Autonomous Challenge}.}
    \label{fig:minerva-vs-tum}
\end{figure}


For the correct planning and execution of overtaking maneuvers, accurate and consistent tracking of the opponent vehicle state and future trajectory prediction are necessary \cite{leon2021review}. At the same time, due to the high velocities involved, reducing the signal processing time and latency is fundamental to react promptly to changes in the environment.

A low-latency, robust, and computationally efficient target tracking algorithm with a high detection range is fundamental for safe and successful autonomous overtaking maneuvers. This paper presents the LiDAR-based vehicle detection and target tracking algorithm deployed on Team PoliMOVE's Dallara AV21 ``MinerVa'' which won first place in all three multi-vehicle IAC competitions it entered.

In this work, we build an online algorithm for reliable vehicle detection from Point Cloud data with a latency lower than the sensor refresh rate, together with the capability of fully observing the target's 2D pose, tracking its motion and estimating its linear and angular velocities, without the availability of labeled data. 


To fulfill these requirements, we implemented a novel Point Cloud segmentation algorithm capable of processing in parallel the three LiDAR sensors mounted on the vehicle, a multi-hypothesis L-shape fitting technique for a racing vehicle moving on a racetrack and a Multi-Target Tracking (MTT) module, which estimates the target speed, heading and yaw rate from position measurements.

The output of this pipeline is then fed to the opponent trajectory prediction and Ego-vehicle trajectory planning module to initiate eventual overtaking or defensive maneuvers.

The main contributions of this paper lie in the following:
\begin{itemize}
    \item A fast and efficient Point Cloud segmentation algorithm working on unstructured Point Clouds with non-uniform, time-varying scan pattern with no information loss.
    \item A 2D pose estimation technique for an irregularly shaped vehicle from a Point Cloud acquired on a racetrack.
    \item A variable-step target tracking algorithm.
\end{itemize}

All of the methodologies presented in this paper are capable of online operation as they have been tested experimentally online by performing fully autonomous overtakes on vehicles travelling at velocities superior to 250 $km/h$.

These contributions are presented in the paper according to the following structure: In Section \ref*{sec: related_work}, we summarize the main research contributions and open challenges in the field of vehicle detection and tracking for autonomous driving. In Section \ref{sec: experimental_setup}, we describe the experimental vehicle used for this research work, while in Section \ref*{sec: methodologies}, we provide a top-down description of all the components of the algorithm we developed. Then in Section \ref*{sec: experimental_results}, we provide a quantitative performance evaluation of every algorithmic step, together with the description of the dataset used for the analysis. Finally, we conclude our work in Section \ref*{sec: conclusions}, by summarizing the approach used and the results achieved while discussing potential future works on this subject.

\section{Related Work}\label{sec: related_work}

\begin{figure}[b]
\centering
    \includegraphics[width=\columnwidth,trim=110 110 185 110]{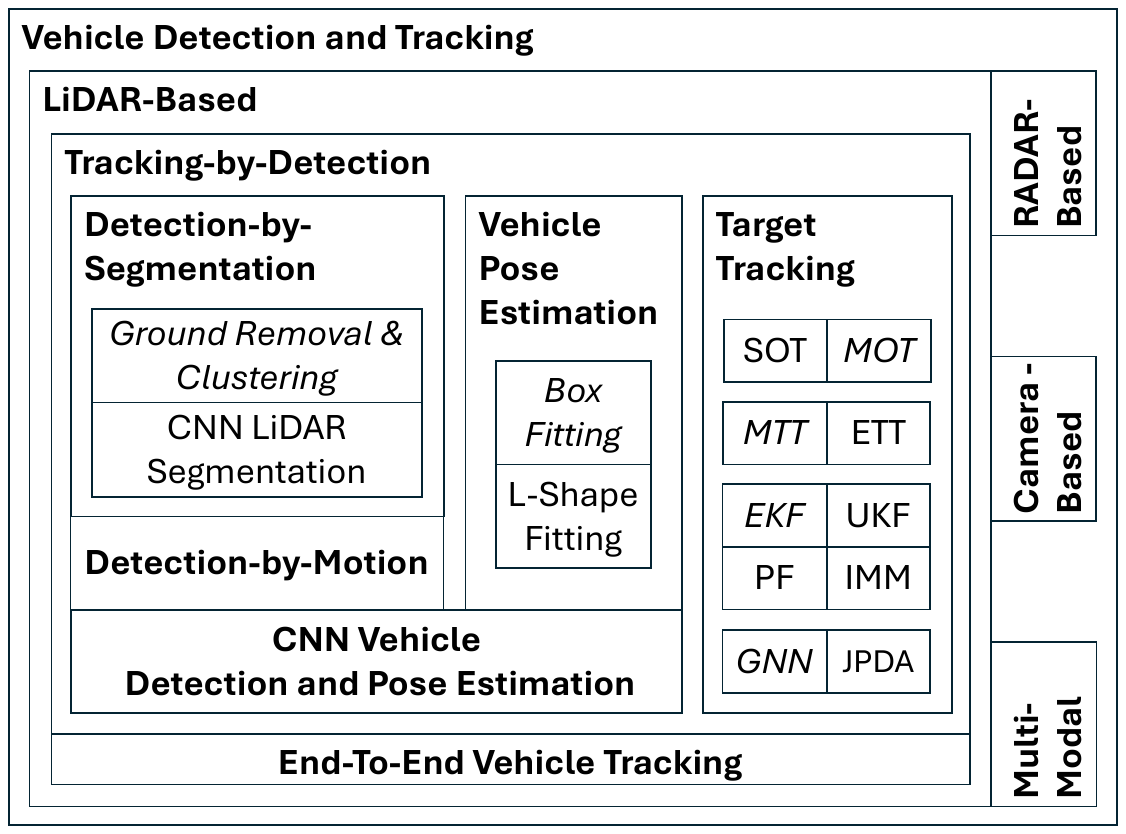}
    \caption{Representation of the main research problems in Vehicle Detection and Tracking, the most common solutions in literature and their relationship. In bold, the research problems, in italic the approaches used in this work. Neighboring boxes should be considered as alternative methodologies.}
    \label{fig: literature}
\end{figure}

This section provides a literature review of the state of the art methodologies for Vehicle Detection and Tracking, with a strong focus on using LiDAR sensors. We will divide the main problem into its principal sub-problems and analyze the main approaches used to solve them.

Figure \ref{fig: literature} shows a taxonomy of the main research problems related to Vehicle Detection and Tracking \cite{wang2023multi} \cite{mao20233d}. In this work, we will focus on LiDAR-based methods, which, despite lacking the velocity measurement and weather robustness of RADARs and the high range and color information of Cameras, provide the greatest position accuracy of the kind, which is crucial for close racing applications. Multi-modal fusion methods, which increase the detection and tracking performance by combining the benefits of the different sensor types, are beyond the scope of this work.

Concerning LiDAR-Based Vehicle Detection and Tracking, the dominant approach in literature is Tracking-by-Detection, which divides Vehicle Detection and Target Tracking as separate problems, as opposed to End-to-End tracking, which extract tracks directly from LiDAR Point Clouds, usually with the use of Convolutional Neural Networks (CNNs). Although in recent years there has been an increase of research in End-to-End tracking as in \cite{pang2021model}, \cite{wang2021ditnet} and \cite{fang20203d}, they are not established methods and have yet to prove their robustness and reliability for online use.

A similar division between algorithmic and data-driven approaches lies in the Vehicle Detection and Pose Estimation problem: although historically treated as separate problems, there has been a growing interest in scientific literature towards CNN-based Vehicle Detection networks, like \cite{lang2019pointpillars}, \cite{shi2020pv} and other approaches mentioned in \cite{alaba2022survey}. 

The development of these data-driven algorithm relies on public, labeled datasets such as KITTI \cite{geiger2013vision}, NuScenes \cite{caesar2020nuscenes}, and Waymo \cite{sun2020scalability}. The vehicles and environments constituting these datasets, however, have very few similarities with our application. To address this gap, the RACECAR dataset \cite{kulkarni2023racecar} has been published by leveraging the contributions of multiple IAC teams, although no labelled version is available at the moment of writing.

Furthermore, the poor generalization capabilities of most data-driven perception algorithms may lead to unsatisfactory behaviour in many edge cases often encountered in racing, like the presence of debris on the track, or a vehicle emitting smoke or spinning out of control. Sharing the considerations expressed in \cite{sauerbeck2023learn}, we decided that data-driven methods were not suitable for this application, as they are difficult to transfer from synthetic data to real life. 

A separate distinction in Tracking-by-Detection methods lies in the detection mechanism: In Detection-by-Segmentation approaches, vehicles are isolated from the surrounding environment by grouping points in the LiDAR point cloud based on geometric features. In contrast, Detection-by-Motion relies on detecting changes in the position of objects over time, using temporal differences between consecutive point cloud frames to identify moving objects. The latter methods usually rely on a representation of the surrounding environment in terms of occupancy, like the Virtual Scan of \cite{petrovskaya2009model} or the Octree in \cite{azim2012detection}. Although simpler and potentially better performing in cluttered environments than Detection-by-Segmentation methods, these methods are incompatible with static obstacles, and therefore would pose a safety hazard in the presence of stopped vehicles on the racetrack.

For these reasons, in this work and in the rest of this literature review, we will focus on a Tracking-by-Detection approach, composed by three main algorithmic steps: Point Cloud Segmentation, Vehicle Pose Estimation and Target Tracking. We will delve into each of these sub-problems in the following sub-sections.

\subsection{Point Cloud Segmentation}\label{subsec:Cloud-segmentation}
Point Cloud segmentation is the process of partitioning a set of 3D points into meaningful groups or segments based on their characteristics or spatial relationships. This process usually facilitates subsequent operations such as object classification and vehicle pose estimation. 

A special case of Segmentation is Ground Removal, which refers to the detection of points belonging to the ground plane. This step is often applied as a pre-processing step for the proper Segmentation algorithm.

In literature, the ground plane has been filtered out by processing the Point Cloud in polar coordinates, using the polar binning method proposed in \cite{himmelsbach2010fast} or the height difference method proposed in \cite{petrovskaya2009model}, and later expanded by \cite{bogoslavskyi2016fast} with the addition of a smoothing filter. However, all these methods work only for mechanical LiDAR sensors, and are not suited for modern MEMS scanning LiDARs.

Concerning autonomous racing, a 2.5D grid approach as in \cite{jung2023autonomous} allows for great computational efficiency, as all the conditions are applied to properties of the grid. However, this method lacks robustness to outliers. A more sophisticated approach is presented in \cite{betz2023tum}, which employs a variation of the algorithm presented in \cite{cho2014sloped} to label the ground points. The main limitations of grid-based approaches are the performance degradation at high range and the inability to process multiple Point Clouds in parallel. 

Beyond Ground Removal, the most established Point Cloud Segmentation method is the Euclidean Clustering algorithm provided by the PCL library in \cite{rusu20113d}, whose computational complexity, however, makes it unsuitable for online use with large Point Clouds without heavy input downsampling, and subsequent information loss.

A mixed approach is employed in \cite{himmelsbach2010fast}, with most of the segmentation performed by searching for connected components on a 2D grid, which can be performed efficiently using image processing techniques, and then a 3D voxel grid is built only for those clusters that need additional segmentation. A similar voxel grid approach is used by \cite{douillard2011segmentation}. However, these methods share the information loss problem of similar 2.5D approaches.

Higher-performing methods try to exploit the structure of the LiDAR scan in order to segment the Point Cloud, usually working in a mix of spherical and Cartesian reference frames. It is the case of the Scan-Line Run algorithm presented in \cite{zermas2017fast}, which first segments the points belonging to a single Point Cloud layer and then looks for correspondences between contiguous layers, or of \cite{bogoslavskyi2017efficient}, which employs a Range Image representation of the Point Cloud to perform the Segmentation. These methods however rely heavily on the structured Scan Pattern of rotating LiDAR Sensors.

A combination of a Range Image representation of the Point Cloud and a CNN has been presented in \cite{milioto2019rangenet++}, providing superior performance compared to classical algorithmic methods. A similar approach based on a cylindrical projection, which maintains the 3D information that is lost during the range image projection, is used in \cite{zhu2021cylindrical}. These methods however share the drawbacks expressed in the previous considerations about CNN-based algorithms, making them unsuited for autonomous racing. 

\subsection{Vehicle Pose Estimation}\label{subssec: pose-estimation}

The goal of Vehicle Pose Estimation in autonomous driving is to infer the vehicle position and orientation, as well as its dimensions, from its corresponding LiDAR points. The two prevalent methodologies are L-shape Fitting and Box Fitting. 

L-shape Fitting, originally developed for 2D LiDARs and then expanded to 3D sensors, aims to model the partial observability due to self-occlusions of rectangular shaped vehicles as passenger cars. These methods use either a variation of the Ramer-Douglas-Peucker algorithm \cite{douglas1973algorithms} to find the cluster's primary orientation, as presented in \cite{ye2016object} and in \cite{sualeh2019dynamic}, or a RANSAC-based approach as in \cite{shen2015efficient} and \cite{zhao2021shape}, and then proceed to fit the minimum area rectangle given its principal orientation. Although very computationally efficient, these methods have a strong reliance on the hypothesis of rectangular shape of the tracked vehicles, which is violated in autonomous racing applications, and are not generally robust to outliers.

Box Fitting techniques instead consist in estimating a bounding box around the entire vehicle by means of the minimization of a cost function, like in \cite{zhang2017efficient}, \cite{kim2017shape} and in \cite{kraemer2018lidar}. Although more robust to outliers than L-shape fitting techniques, as they do not rely explicitly on the rectangular shape on the tracked vehicle, they have a high variance in estimation, and they have not been tested with vehicles other than road cars or trucks.

\subsection{Target Tracking}\label{subsec: target-tracking}

Target Tracking is the process of estimating the number of objects of interest (targets) present in the tracking volume, together with their partially measurable states. When applied to Autonomous Ground Vehicles (AGV) usually involves estimating the number of moving vehicles surrounding the EGO vehicle, their 2D pose and linear and angular velocity. 

Considering the taxonomy shown in Figure \ref{fig: literature}, together with the definitions expressed in \cite{blackman1999design}, the first division of Tracking algorithms lies in the difference between Single Object Tracking (SOT), which is employed in single-objective control systems, and Multiple Object Tracking (MOT) techniques, which are more appropriate to autonomous driving applications, where the number of targets is unknown and can be larger than 1.

MOT problems can be further divided into Multi-Target Tracking (MTT) problems, where each target is supposed to generate at most one measurement at each iteration, \cite{vo2015multitarget}, and Extended Target Tracking (ETT) problems, where each target can generate more than one measurement, as it is usually the case with LiDAR sensors. ETT techniques for vehicle tracking are usually based on strong assumptions about the vehicle shape and should guarantee a larger precision in tracking than MTT, although some shape-free ETT techniques exist as in \cite{pieroni2024design}. Although ETT techniques have been successfully applied to autonomous driving \cite{granstrom2014multiple} \cite{dahal2022extended}, their strong assumptions about the rectangular shape of the target make them unsuitable for tracking a racecar.

All Target Tracking algorithms use a dynamical filter in order to estimate the target state and approximate its uncertainty distribution. The most common filters employed are the Extended Kalman Filter (EKF), the Unscented Kalman Filter (UKF), the Particle Filter (PF) and the Interacting Multiple Models (IMM) estimators.

The classical approach to the Multi-Target Tracking problem involves the use of an Extended Kalman Filter (EKF) as in \cite{kim2017shape}, which has also successfully been applied to autonomous racing in \cite{karle2023multi}. Although simple, this filter has proven to be effective in order to plan and execute autonomous overtaking maneuvers at high speed.

A more sophisticated approach involves the use of multiple EKF estimators in the Interacting Multiple Models (IMM) framework. \cite{jung2023autonomous} presents an autonomous racing application of a IMM-UKF-PDAF tracker, although it does not provide a quantitative performance evaluation of the algorithm performance. In \cite{petrovskaya2009model}, the authors employ a PF to track the moving vehicles, but the higher computational load intrinsic in the PF does not scale well with the number of targets in the scene.

The last aspect of Target Tracking algorithms lies in the mechanism for Data Association, which can be a simple Global Nearest Neighbor (GNN) association, based on the minimization of a cost function like in \cite{jo2016tracking} or a more computationally intensive Joint Probabilistic Data Association (JPDA) algorithm as in \cite{sualeh2019dynamic}.

\subsection{Open Challenges}

This section has provided an overview of the key sub-problems involved in the LiDAR-based Vehicle Detection and Tracking for Autonomous racing. Most of the scientific literature on this topic aims to solve a subset of the problems schematized in Figure \ref{fig: literature} under a variety of operational environments, although mostly focused on urban driving.

The most established Point Cloud Segmentation algorithms have large computational latencies that make them unsuitable for online use at high speed without significant information loss due to downsampling.  The few methods designed for high computational efficiency suffer from lack of robustness and generality as they work under very restrctive assumptions. To the best of the authors' knowledge, there is no efficient, general and highly performing Point Cloud Segmentation algorithm available in scientific literature.

Concerning the Vehicle Pose Estimation problem, the scientific literature proposes solutions only under the assumption of the rectangular shape of the vehicles, which is not valid in this application, and will require an ad-hoc solution. On the other hand, Target Tracking methodologies are well-established, allowing us to use state-of-the-art algorithms.

This work targets the research gap concerning fast and efficient Point Cloud Segmentation algorithms with broader domain assumptions then those available in literature, together with a novel solution to the Vehicle Pose Estimation problem applied to Autonomous Racing.

\section{Methodologies}\label{sec: methodologies}

\tikzset{
block/.style = {draw, fill=white, rectangle, minimum height=2em, minimum width=5em},
midblock/.style = {draw, fill=white, rectangle, minimum height=2em, minimum width=12em},
bigblock/.style = {draw, fill=white, rectangle, minimum height=2em, minimum width=16em},
}

\begin{figure}
    \centering
    \begin{tikzpicture}
        \node[block, align=center, rounded corners=5pt] at (0,0) (sx) {LiDAR 1};
        \node[block, align=center, rounded corners=5pt] at (3,0) (fr) {LiDAR ..};
        \node[block, align=center, rounded corners=5pt] at (6,0) (dx) {LiDAR n};
        \node[block, align=center, text width=6em] at (0,-1.5) (sx_s) {Range Image Segmentation};
        \node[block, align=center, text width=6em] at (3,-1.5) (fr_s) {Range Image Segmentation};
        \node[block, align=center, text width=6em] at (6,-1.5) (dx_s) {Range Image Segmentation};
        \node[bigblock, align=center] at (3,-3.0) (meas) {Vehicle Detection and Pose Estimation};
        \node[bigblock, align=center] at (3,-4.25) (trk) {Target Tracking};
        \draw [->] (sx) -- (sx_s);
        \draw [->] (fr) -- (fr_s);
        \draw [->] (dx) -- (dx_s);
        \draw [->] (sx_s) -- (meas);
        \draw [->] (fr_s) -- (meas);
        \draw [->] (dx_s) -- (meas);
        \draw [->] (meas) -- (trk);
    \end{tikzpicture}
    \caption{Block scheme of the LiDAR-Based Tracking algorithm}
    \label{fig:block-scheme-algorithm}
\end{figure}
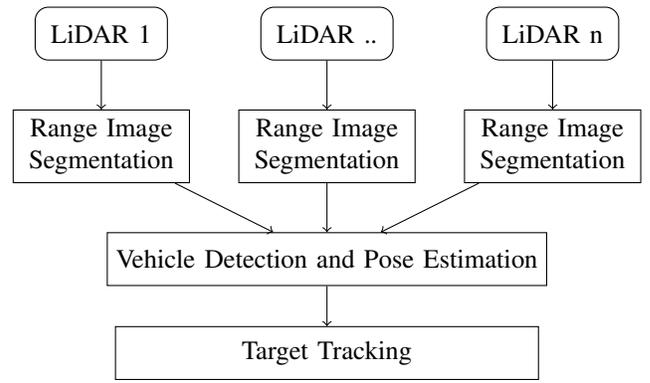

In this section, we present our approach to solving the problem of LiDAR-Based Vehicle Detection and Tracking for Autonomous Racing. We first describe the main assumptions and algorithmic structure, and then we delve into more detail concerning the main algorithmic steps. 

We propose an algorithm that operates under the following domain assumptions:

\begin{itemize}
    \item The LiDAR sensors provide an unstructured Point Cloud, with a potentially non-uniform and time-varying scan pattern.
    \item The LiDAR sensors provide, other than the XYZ coordinates, Intensity and Ring information for each point.
    \item The platform (EGO) vehicle is equipped with a number of LiDAR sensors with adjacent or overlapping Field of View.
    \item The size of the vehicles to be tracked is a-priori known.
    \item Precise EGO vehicle absolute localization is available, together with a map of the operative environment.
\end{itemize}

Figure \ref*{fig:block-scheme-algorithm} shows the algorithmic architecture of this work. The process begins with Point Cloud Segmentation, where we apply a Range Image-based Ground Removal and Clustering algorithm. This process is performed in parallel on the three input Point Clouds. We then merge the three Segmented Point Clouds, discarding clusters with size and position incompatible with a racecar. Next, we estimate the 2D pose of the vehicles using a robust method based on the information from the track layout. Finally, the Target Tracking algorithm tracks the opponent vehicles, predicts their future states, and estimates their velocities. The Target Tracking output is fed to the Local Planning module.

The problem of motion distortion in LiDAR scans, which is of great importance for mapping applications and whose effect is particularly strong in this application due to the high EGO vehicle speed, is not relevant to the Vehicle Detection and Tracking due to the small relative speed between the vehicles, and therefore was not addressed in this work.

\subsection{Point Cloud Segmentation}

\begin{figure}
    \centering
    \def\leftcropping{-28}
    \def\rightcropping{0}
    \def\scaledheight{0.08\linewidth}

    \subfloat[Input Point Cloud $\mathbf{P}$ in spherical coordinates (blue: near, red: far) \label{subfig:pt-cloud-lum}]{\includegraphics[trim={0, 0, \rightcropping, 0},width=\linewidth,clip]{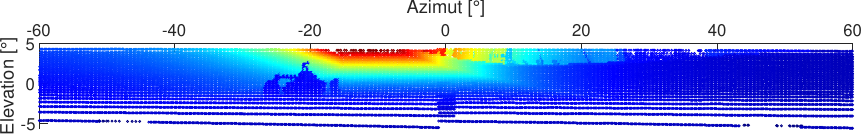}}

    \subfloat[Range Image $\mathbf{R}$ (blue: near, red: far) \label{subfig:range-image-lum}]{\includegraphics[trim={\leftcropping, 0, \rightcropping, 0},width=\linewidth,clip,height=\scaledheight]{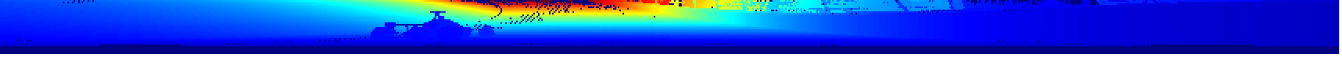}}
    \\
    
    \subfloat[Angle Image $\mathbf{A}$ (blue: horizontal, red: vertical)\label{subf: angle-image}]{\includegraphics[trim={\leftcropping, 0, \rightcropping, 0},width=\linewidth,clip,height=\scaledheight]{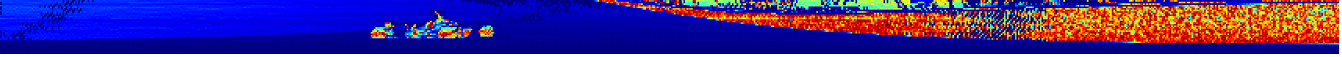}}
    \\

    \subfloat[Smoothed Angle Image $\mathbf{\hat{A}}$ (blue: horizontal, red: vertical)\label{subf: smooth-angle}]{\includegraphics[trim={\leftcropping, 0, \rightcropping, 0},width=\linewidth,clip,height=\scaledheight]{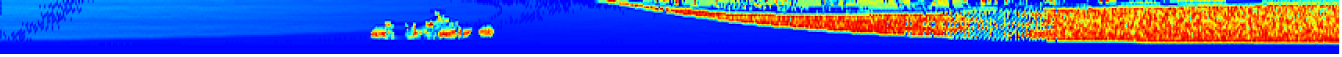}}
    \\

    \subfloat[Repaired non-Ground Range Image $\mathbf{\hat{R}}$ (blue: near, red: far)]{\includegraphics[trim={\leftcropping, 0, \rightcropping, 0},width=\linewidth,clip,height=\scaledheight]{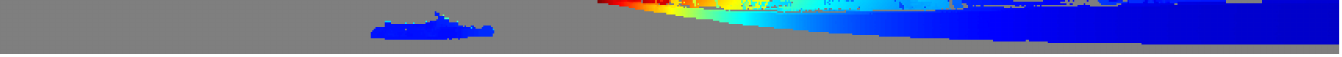}}
    \\
    
    \subfloat[Label Image\label{subf:label-image} $\mathbf{L}$ (color: label)]{\includegraphics[trim={\leftcropping, 0, \rightcropping, 0},width=\linewidth,clip,height=\scaledheight]{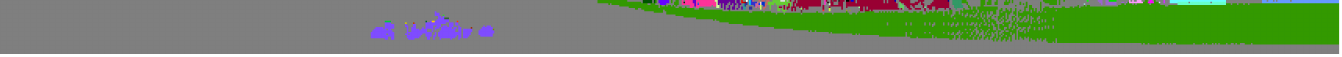}}
    \\

    \subfloat[Segmented Point Cloud (color: label)\label{subf:segmented-cloud}]{\includegraphics[trim={20, 20, 20, 20},width=\linewidth,clip]{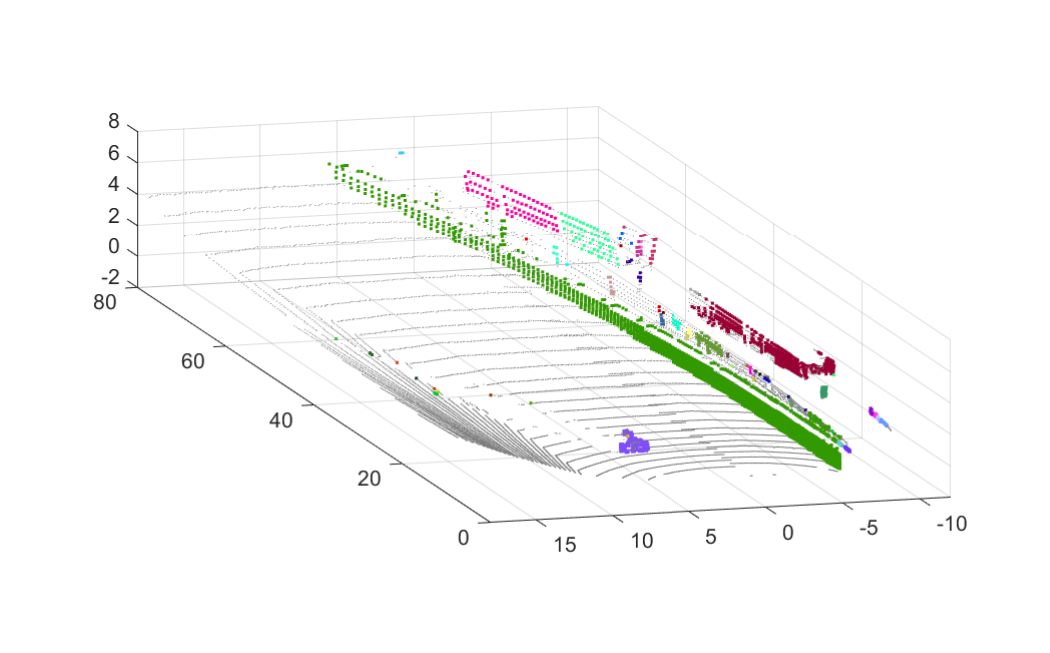}}
    \\
    
    \caption{The five steps of the Point Cloud segmentation algorithm and the resulting segmented Point Cloud. Images are scaled vertically for easier visualization.}
    \label{fig:met-range-angle-label image}
\end{figure}

The Point Cloud segmentation algorithm presented in this work is an expansion of the Range Image Clustering method presented in \cite{bogoslavskyi2016fast}. A Range Image $\mathbf{R}$ is a 2D dense representation of a 3D Point Cloud $\mathbf{P}$ obtained by transforming $\mathbf{P}$ into spherical coordinates and projecting it on the azimuth-elevation plane. It comes in the form of a $n\times m$ matrix where each row $i\in[1,n]$ represents a constant elevation value and each column $j\in[1,m]$ represents a constant azimuth value. The element $r_{i,j}\in\mathbf{R}$ represents the measured range for a given azimuth and elevation, such that the Range Image representation of a point is equivalent to its 3D coordinates, as $\forall p \in \mathbb{R}^3, p \subset \mathbf{P} , p_{k} \leftrightarrow r_{i,j}$.

This representation works well with a rotating LiDAR. A rotating LiDAR allows for easy construction of the Range Image as the distribution of the scan points on the azimuth-elevation plane is fixed, as it depends on the geometrical characteristics of the sensor. This is not the case for a scanning LiDAR:  due to the simultaneous horizontal and vertical motion of the sensor head, the elevation is not constant along the scan lines, as Figure \ref{subfig:pt-cloud-lum} shows. Moreover, some sensor models allow for online configuration of the scan pattern, making this azimuth-elevation characteristic time-varying. 

If the sensor provides the information about which scan layer originated every point, however, it is possible to construct a range image $\mathbf{R}$ on the azimuth-layer plane, where each row $l\in[1,n]$ corresponds to a single scan line and every element $r_{l,j}\in\mathbf{R}$ represents the measured range of the scan point belonging to layer $l$ and azimuth corresponging to column $j$. An example of such construction is presented in Figure \ref{subfig:range-image-lum}. 

This azimuth-layer Range Image is not sufficient however to reconstruct the original image, as it is lacking information about the elevation of the points. This information is stored into an Elevation Image $\mathbf{E}$, a second matrix of the same size as $\mathbf{R}$, where each element $e_{l,j}\in\mathbf{E}$ contains the elevation value of the point whose range is stored in $r_{l,j}$. In this way, it is possible to define the mapping $p_{l,j} \leftrightarrow \{r_{l,j},e_{l,j}\}$  to reconstruct the original $\mathbb{R}^3$ point without loss of information.

Many LiDAR sensors have the capability of providing multiple range readings for every scan line. However, the value of every pixel of $\mathbf{R}$ must be unique (or empty). Thus, a policy is needed to determine which point to retain. For this application, we chose to keep the point with the largest intensity value, as it is less likely to belong to dust or debris. 

On the other hand, $\mathbf{R}$ may figure several empty pixels, as Figure \ref*{subfig:range-image-lum} shows, caused by missed scan returns especially at large distances, or an uneven spacing across the azimuth. These points are artificially filled using a mean filter across the non-empty neighbors in the same image column.

After completing the construction of $\mathbf{R}$ and $\mathbf{E}$, following the approach presented in \cite{bogoslavskyi2017efficient}, we build the Angle Image $\mathbf{A}$. It consists of a $n-1\times m$ matrix, containing the angle between the $x-y$ plane (in the sensor reference frame) and the vector connecting two points with the same azimuth belonging to consecutive layers, as in  \eqref{eq: angle-image}.

\begin{equation}
    a_{i,j}\in\mathbf{A} = \arctan \left(\frac{ r_{i,j}\cos(e_{i,j}) - r_{i+1,j}\cos(e_{i+1,j})}{ r_{i,j}\sin(e_{i,j}) - r_{i+1,j}\sin(e_{i+1,j})} \right)
    \label{eq: angle-image}
\end{equation}

Figure \ref*{subf: angle-image} shows the result of the computation of the Angle Image $\mathbf{A}$. Then, we smooth it by applying a column-wise Savitzky-Golay kernel with window size $sz_{SG}$ to $\mathbf{A}$, obtaining the Smoothed Angle Image $\mathbf{\hat A}$ which is shown in Figure \ref*{subf: smooth-angle}. By imposing a threshold $th_{gnd}$ over the maximum allowed ground slope, we can label as \textit{not ground} the points where $\hat{a}_{i,j} > th_{gnd}$. Since $\mathbf{A}/\mathbf{\hat A}$ has one row less than $\mathbf{R}$, the bottom row is by default labelled as \textit{ground}.

The resulting non-ground Range Image is then fed to the angle-based segmentation step introduced in \cite{bogoslavskyi2016fast}, where every connected component of $\mathbf{R}$ is further segmented via the Breadth-First Search algorithm that compares every element of  with its 4-neighbors. In particular, two points $p_1,p_2\in P$ being neighbors in $\mathbf{R}$ are allowed in the same cluster $ {C}$ if the angle $\beta_{1,2}$, representing the incidence of the segment connecting the two points with the segment connecting the first and the origin, is lower than a threshold $th_{seg}$. The angle $\beta_{1,2}$ is defined in \eqref{eq: beta}

\begin{equation}
    \beta_{1,2} = \arctan \left( \frac{r_1\sin\alpha_{1,2}}{(r_1-r_2)\cos\alpha_{1,2}} \right)
    \label{eq: beta}
\end{equation}
where $\alpha_{1,2}$ represents either the azimuth difference of a pair row neighbors or the elevation difference of a pair of column neighbors. The output of this algorithm is a Label Image $\mathbf{L}$ whose element $l{i,j}\in\mathbf{L}$ represents the univocal index of the cluster associated to the point $p_{k}\leftrightarrow r_{i,j}$

A strong limitation of the approach described in \cite{bogoslavskyi2016fast} lies in its inability to effectively manage a cluster that extends across multiple connected components within the range image, as the Breadth-First-Search algorithm is applied only to elements belonging to the same connected component of the non-ground Range Image.

Usually, the front and rear suspension brackets of the opponent vehicle, composed of small and opaque carbon fiber pipes, are either not directly hit by a LiDAR ray or they provide no reflection, being black. At the same time, the cars presents multiple horizontal surfaces that can be labeled as ground, like the front and rear wings and the under-body. This usually results in having multiple separated non-ground components, as shown in Figures \ref{subf: angle-image} and \ref{subf: smooth-angle}.

Therefore, we performed the Non Ground Range Image Reparation (NGRIR) after the Ground Removal step, by applying a horizontal mean filter kernel of size $w_s$ to the \textit{ground} elements of $\mathbf{R}$, iterating only over the non-ground elements in the same row of the range image. An additional check over the maximum depth difference $th_r$ of the symmetric element pairs avoids the merging of connected components belonging to different elements, as expressed in \eqref{eq: NGRIR}. 

\begin{equation}\label{eq: NGRIR}
    \begin{aligned} 
        N(r_{i,j})  &= \{r_{i,j+k}, k \in [-w_s,+w_s]   &   \\
                    & \quad \quad \quad \| \exists r_{i,k} \wedge  |r_{j-k}-r_{j+k}| < th_r \}  \\
        r^*_{i,j}   &= \sum N(r_{i,j}) / |N(r_{i,j})|  & \\
    \end{aligned}
\end{equation}

The result of the NGRIR algorithm is the Reparied Non-Ground Range Image $\mathbf{\hat{R}}$ shown in Figure \ref{subf:label-image}, while Figure \ref{subf:segmented-cloud} shows how this technique allows the Segmentation Algorithm to assign the same label to a cluster spanning multiple connected components in the non-ground range image.

So far, the input Point Clouds coming from the multiple LiDAR sensors have been processed in parallel on different CPU cores. When the Segmentation step is terminated, the algorithm collects the segmented Point Clouds and applies a 3D rigid transformation to bring them to a common, vehicle-fixed reference frame. Then, we perform Cluster Merging in order to obtain a single, segmented Point Cloud.

To determine which pairs of clusters $C_1$ and $C_2$ from two neighboring Point Clouds $\mathbf{P_1}$ and $\mathbf{P_2}$ belong to the same object and therefore should be merged, we apply the following equation
\begin{equation}
    C_1 \equiv C_2 \quad \text{if} \quad \exists i,j\in[1,n] \quad \| \quad |\mathbf{p^1_{i,m}} - \mathbf{p^2_{j,1}}| < {th_{mrg}} 
    \label{eq: cloud-merging}
\end{equation}
to the clusters belonging to the last column of $\mathbf{R_1}$ and the first column of $\mathbf{R_2}$ (for example, the last/right column of the Range Image corresponding to the Front LiDAR scan and the first/left column of the Range Image corresponding to the Right LiDAR scan), where ${th_{mrg}}$ is the euclidean distance threshold to consider the two points part of the same cluster. Although simple, this brute force approach works well in practice, and it proved to be robust to small errors in extrinsic sensor pose calibration. 

\begin{figure}
    \centering
    \def\topcropping{240}
    \def\bottomcropping{280}
    \def\leftcropping{200}
    \def\rightcropping{380}
    
    \subfloat[Input Point Clouds with intensity measurements (red: lowest, blue: highest)]{\includegraphics[trim={\leftcropping, \bottomcropping, \rightcropping, \topcropping},width=\linewidth,clip]{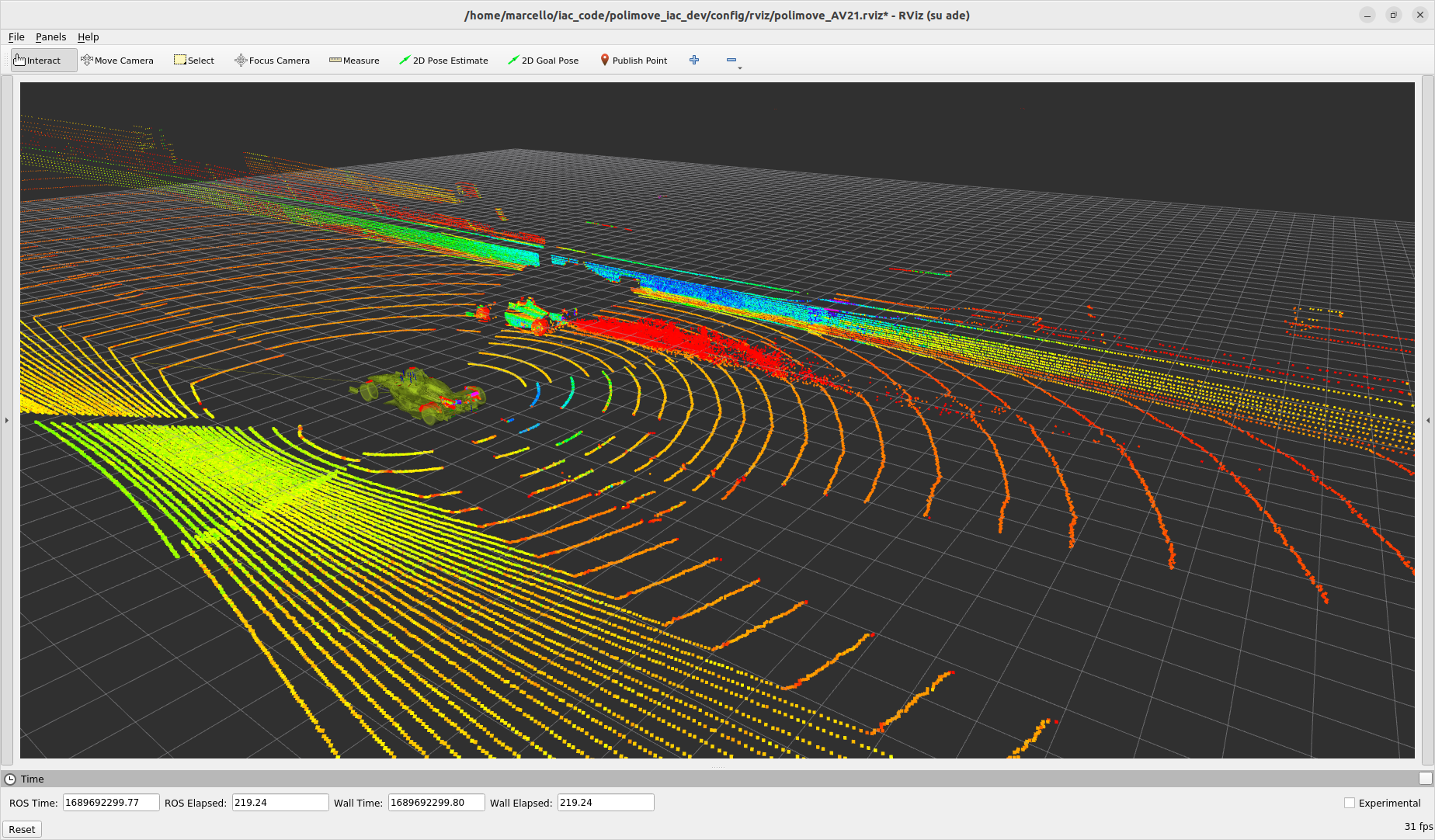}}
    
    \subfloat[Segmented merged Point Cloud (color: label)]{\includegraphics[trim={\leftcropping, \bottomcropping, \rightcropping, \topcropping},width=\linewidth,clip]{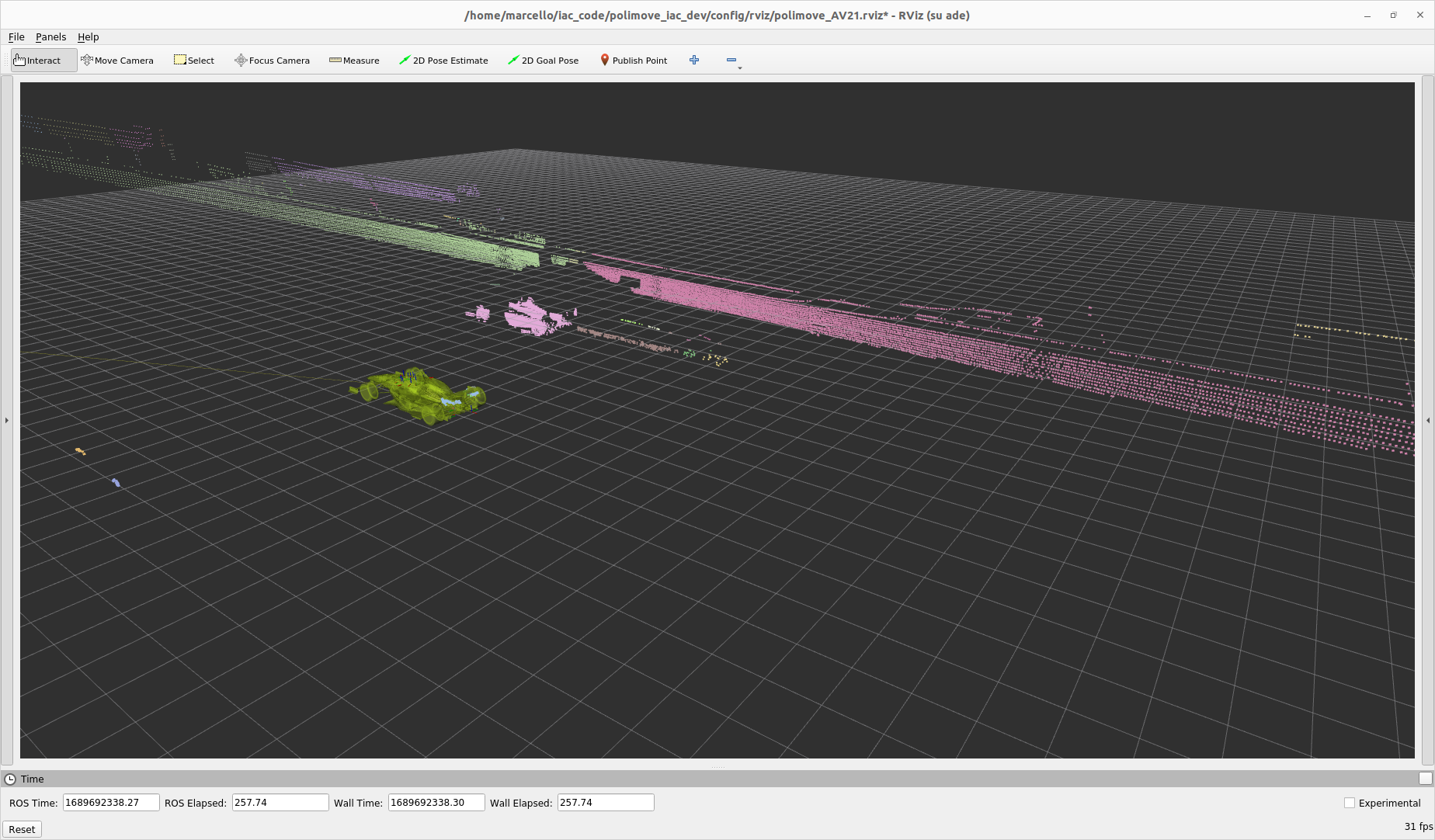}}
    
    \caption{Input and Segmented Point Clouds from the three sensors capturing an opponent performing an overtaking maneuver at 250 km/h. The EGO vehicle is represented as a semi-transparent 3D graphical model. Grid size: 1m.}
    \label{fig: raw_vs_segmented}
\end{figure}

Figure \ref{fig: raw_vs_segmented} shows the qualitative results of the cluster merging process. The front wheel captured by the frontal LiDAR is correctly merged with the rest of the opponent vehicle captured by the right sensor. At the same time, the segments of the wall captured in the front scan are joined with their corresponding ones from the right LiDAR, but the cluster is immediately broken as the presence of the opponent car does not allow the wall to form a single connected component in the Range Image of the right Cloud.

\subsection{Vehicle Detection and Pose Estimation}

After merging the clusters across different sensors, we discard all the clusters with the longest dimension larger than the a-priori known vehicle size. Furthermore, we use the EGO vehicle pose estimation to project every cluster centroid on a global map of the track surface in order to discard all the objects lying outside the map. 

Although the use of the map to classify clusters adds a risk of false negative detections in case of an imprecise EGO pose estimation, this step is fundamental to filter out false detections originating from large overhanging signs or vehicles parked in the pit lane, which do not partecipate in the on-track action.

For every cluster labelled as a vehicle, we want to estimate its 2D pose $\{x,y,\psi\}_{opp}$ from the 3D scan points. We will follow the most common method to solve this problem by fitting a rectangular model to the points.

A rectangle is defined by 5 variables: $\{x,y,\psi,W,L\}$, however the most important is the principal heading $\psi$ as, given a certain heading, finding the minimum area rectangle enclosing all the cluster points is trivial. Due to occlusions in the LiDAR scan, however, the width and length of the minimum area rectangle are smaller than the actual vehicle dimensions. For this application, the width and length of the vehicle are a-priori known, and can be imposed as constraints to the rectangle fitting problem.

Therefore, given a set of points belonging to a cluster $\{p_i\}\subseteq\mathbf{C}$, and a estimated heading $\psi_{opp}$, it is immediate to find the coordinates of the corner point $p_c = \{x,y\}_c$ by fitting the minimum area rectangle with principal heading $\psi$. Once the corner point has been found, it is possible to impose the known vehicle width $W=W_{EGO}$ and length $L=L_{EGO}$, and to compute the coordinates of the vehicle geometrical center $\{x,y\}_{opp}$ in the EGO-centered reference frame. 

We propose an innovative method for estimating $\psi_{opp}$ which is based on the fusion of two independent estimators: a trajectory-based heading estimation, $\hat\psi_{REF}$ and a classical L-shape fitting technique based on the bidimensional rectangle fitting error variance minimization $\hat\psi_{VAR}$.

The first method computes $\hat\psi_{REF}$ using the prior knowledge of the track map by assuming the opponent to be traveling parallel to the center line. Given the coordinates of the cluster centroids transformed in the map-fixed reference frame, it is trivial to project it over the centerline and compute its heading, which once tranformed back in the vehicle-fixed reference frame becomes $\hat\psi_{REF}$. The rationale behind this estimator lies in the fact that in high speed oval racing the vehicles tend to have negligible incidence angles with respect to the track centerline, as the sideslip angles are reduced and overtake or avoidance maneuvers require little heading changes.

This method represents an open-loop estimation which, is not dependent on the real opponent vehicle heading. While the approximation $\psi_{opp} \approx\hat\psi_{REF}$ holds well under nominal racing conditions, where it can be more robust than actual L-shape estimators, it clearly fails under non-nominal conditions like a vehicle spinning without control or stopped sideways on the track surface.

The second estimator $\hat\psi_{VAR}$ aims to estimate directly the minimum area rectangle from the 3D Point Cloud following the Variance Minimization method described in \cite{zhang2017efficient}.

The fusion of the two methods happens according to \eqref{eq: psi_hat}

\begin{equation}
    \hat\psi_{opp} = \begin{cases}
        \hat\psi_{REF} & \text{if} \quad \lvert P_{REF}\rvert \geq \lvert P_{VAR} \rvert \\
        \hat\psi_{VAR} & \text{otherwise}
    \end{cases}
    \label{eq: psi_hat}
\end{equation}
where $P_{REF}$ and $P_{VAR}$ indicate respectively the cardinality of the set of points inside the estimated rectangle using $\hat\psi_{REF}$ and $\hat\psi_{VAR}$, respectively.

In this way, we are certain to always choose the most representative estimation of the opponent heading, with a preference towards the open-loop estimator when no substantial difference is present. 

\begin{figure}
    \centering
    \includegraphics[trim={0, 0, 0, 0},width=\linewidth,clip]{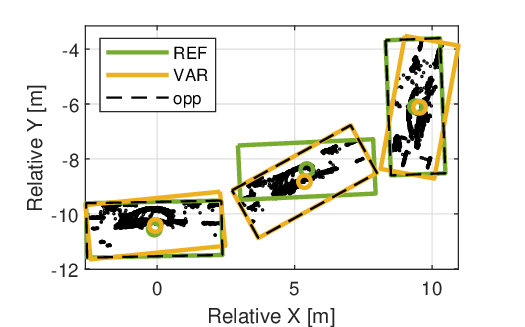}
    \caption{Graphical comparison of the results of the two rectangle fitting algorithms $\hat\psi_{REF}$ and $\hat\psi_{VAR}$ over the segmented Point Cloud of a vehicle entering a spin.}
    \label{fig:l-shape}
\end{figure}

Figure \ref*{fig:l-shape} shows qualitatively the results of the double estimator: while in the first and third Cloud the difference between $\hat\psi_{REF}$ and $\hat\psi_{HAT}$ is low, and the two estimators are equivalent, in the middle Point Cloud the most correct estimation comes from $\hat\psi_{HAT}$, and it is chosen as $\lvert P_{REF}\rvert \geq \lvert P_{VAR} \rvert$.

\subsection{Target Tracking} 

To track the opponent vehicles, estimate their velocity, and predict their future state, we employ a Multi-Target Tracking algorithm that uses a variable-rate Extended Kalman Filter to estimate the target states with Global Nearest Neighbor data association and a three-state M/N track management logic. For a more comprehensive dissertation on Target Tracking please refer to \cite{blackman1999design}.

The variable-step EKF update happens either at the end of the pose estimation algorithm or after a certain time since the last received measure. In case of new measurements, the time step is computed by sensor timestamp subtraction, to be robust to differences in processing latency between consequent scans. Before publishing the output, the filter performs one last prediction step to provide the most up-to-date estimation of the target state.

The model used for the tracking EKF is a Constant Velocity and Turn Rate (CVTR), also known as the Coordinated Turn (CT)  model. This model is described the following equation:
\begin{equation}\label{eq: CVTR}
    \\
    \dot{X}(t) = 
    \begin{cases}
    \dot{x}(t)       = v(t)\cos(\phi(t)) \\
    \dot{y}(t)       = v(t)\sin(\phi(t)) \\
    \dot{v}(t)       = 0 \\
    \dot{\phi}(t)    = \omega(t) \\
    \dot{\omega}(t)  = 0 
    \end{cases}   
\end{equation}
and it represents the motion of a rigid body on a plane, defined by its 2D pose $\{x,y,\phi\}$ in the map-fixed Cartesian reference frame with constant linear and angular velocity $v$ and $\omega$.

The measurements used for the filter update are the $x$ and $y$ centroid position transformed in the map-fixed reference frame using the EGO vehicle pose at the instant of the LiDAR frame. For this work, we decided not to include the measured heading $\psi$ as a filter input. The rationale behind this choice is that in critical conditions, like a vehicle spinning out of control, the course angle $\phi$ and the vehicle heading $\psi$ diverge due to the high sideslip angle. To properly handle an unreliable measure in an EKF is beyond the scope of this work.

The measurement model is linear in the state space and it is described by
\begin{equation}\label{eq: EKF_output}
    \begin{aligned}
    Y(t)    & = HX(t) \\
    H       & = \begin{bmatrix}
                1 & 0 & 0 & 0 & 0 \\
                0 & 1 & 0 & 0 & 0
                \end{bmatrix}
    \end{aligned}
\end{equation}

The Data Association step uses the Mahalanobis Distance between the measured and tracked position weighted by the EKF output error covariance matrix $\mathbf{S}$. A gating threshold is imposed over the distance to avoid wrong associations, defining an association region for every track. Once the distances have been computed for all the feasible association pairs, the Munkres algorithm is used in order to determine the best assignment.

The track life cycle is determined by a two threshold \textit{M/N logic}: every measure that is not associated with a track will create a new \textit{tentative} track. The track state will transition to \textit{confirmed} once a minimum number $M_c$ of associations have taken place in the last $N$ iterations. In contrast, the track will become \textit{dead} if less than $M_d$ iterations have taken place in the last $N$ iterations.

\section{Experimental Setup}\label{sec: experimental_setup}

\begin{figure}
    \centering
    \includegraphics[angle = 90,width = \linewidth]{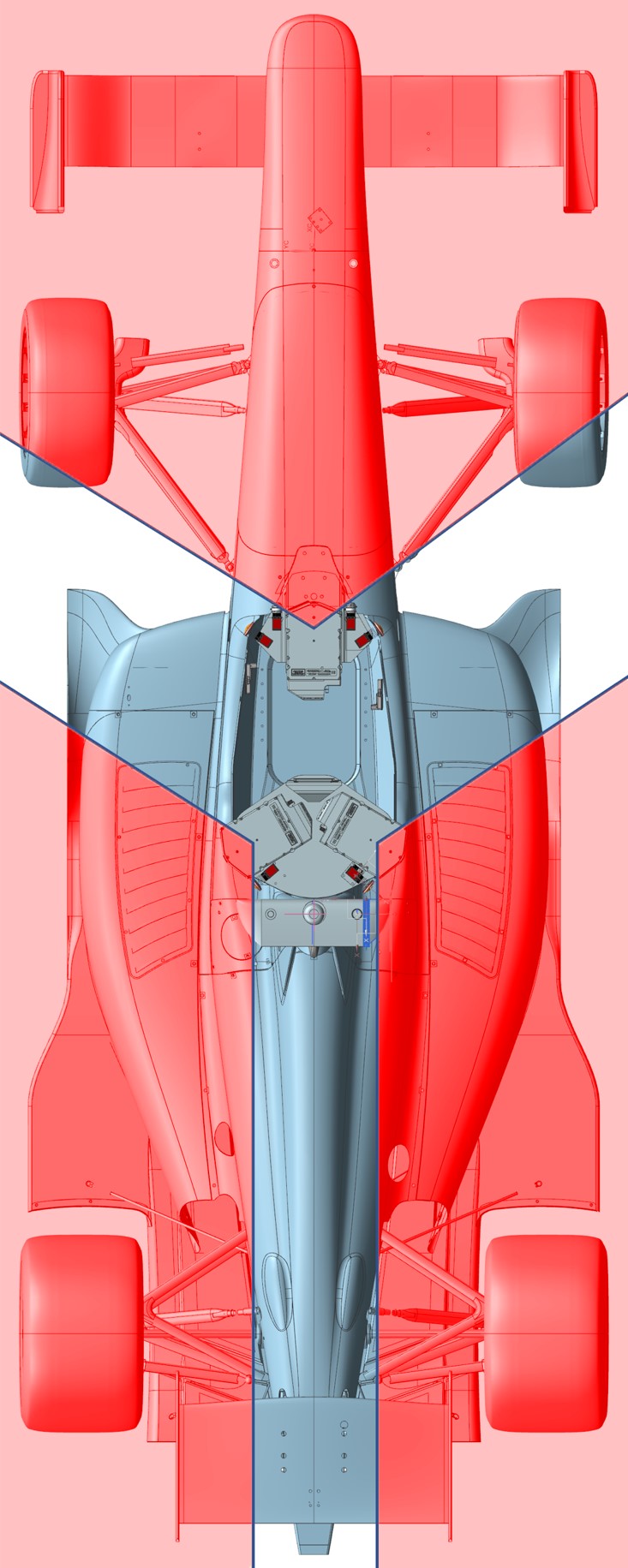}
    \caption{LiDAR sensors mounting positions and horizontal FOV coverage of the three sensors}
    \label{fig:lidar-mounting-position}
\end{figure}

The platform used for this research is the Dallara AV21 shown in Figure \ref{fig:minerva-vs-tum}, which is a level 4 autonomous vehicle based on the single-seat Dallara IL-15 chassis, with additional sensors and actuators.

This vehicle is capable of speeds up to 310 $km/h$ in an oval circuit configuration, with 18.5 and 26.5 $m/s^2$ (2 and 2.7 $g$) maximum longitudinal and lateral acceleration, respectively. As the Dallara AV21 is the only vehicle admitted to the IAC competitions, these performance bounds are shared by both the ego and tracked vehicles.

The vehicles are provided with a redundant dual-antenna RTK GNSS + IMU sensor, which allows for accurate online positioning and state estimation of the ego-vehicle, while the opponent vehicle data provide precise ground truth for offline performance analysis. The onboard computer is a dSPACE Autera \cite{AUTERA} with a 12-core CPU working at a frequency of 2.0 GHz and 128 GB of 2133 MHz RAM. It runs the Ubuntu 20.04 operating system, as our software is written in C++ using the ROS Galactic \cite{macenski2022robot} middleware.

The car is equipped with three Luminar Hydra LiDAR sensors, which are the main subject of this work. The Luminar Hydra sensor \cite{Hydra} is a vertically scanning LiDAR with a fixed 120° horizontal Field-Of-View (FOV) and a software configurable vertical FOV in a subset of the {[-15°,+15°]} range. 

Figure \ref{fig:lidar-mounting-position} shows the mounting positions of the three sensors around the cockpit of the vehicle. This configuration results in a 40 cm wide blind spot between the front and lateral sensors. The presence of the rear wing and its end plates creates a noticeable blind spot in the rearward direction, observable in Figure \ref{fig: raw_vs_segmented}, which severely impacts the opponent detection performances in that area.

Each sensor outputs 640 scan lines per second, each one composed of around 850 points, for a total of more than 500,000 points per sensor per second. The sensor scan rate can also be configured online in the [1,30] Hz range. We found a good compromise between latency and scan density by configuring our sensor to operate at a fixed 20 Hz frequency, resulting in 32 scan lines per Point Cloud. 

Every scan point is characterized, other than its cartesian coordinates, by its scan layer index, an intensity measurement, and its individual timestamp. The vertical FOV of the sensor, together with the distribution of the scan lines, can be configured online and therefore time-varying.

\section{Results}\label{sec: experimental_results}

In this section, we will perform a quantitative analysis of the performance of every algorithm module. We will evaluate both the computational complexity and a set of specific metrics for every algorithmic step: Point Cloud Segmentation, Vehicle Detection, Vehicle Pose Estimation and Target Tracking. 

We will evaluate the algorithms performance over a dataset acquired during the 2023 IAC @ CES competition, which took place on January 7, 2023, at the Las Vegas Motor Speedway, using the parameters described in Table \ref{tab: params}. The dataset contains a total of 15 overtake maneuvers with opponent speeds ranging from 160 km/h to 250 km/h and EGO speeds up to 278 km/h, captured both from the overtaking and overtaken vehicle perspective.

\begin{figure}[t]
    \centering
    \includegraphics{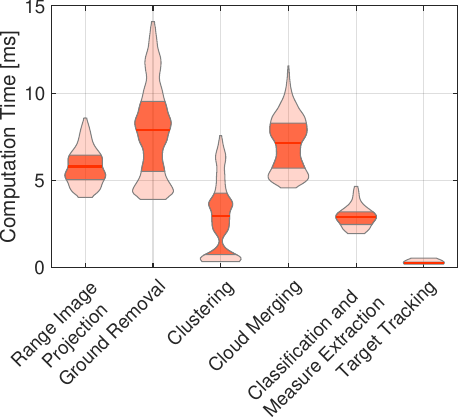}
    \caption{Violin plots of the computation time of the main modules of the proposed method on the 2.0 GHz Intel Xeon CPU of the onboard computer. The distributions are cropped to the 5th and 95th percentile, with the 25th and 75th percentiles and the mean value highlighted}
    \label{fig:res-computation-times}
\end{figure}

Figure \ref{fig:res-computation-times} shows the distribution of the computation times of the main algorithmic modules on the onboard computer during the race. The highly optimized nature of the algorithms allows for an average processing delay of 26 ms, which is half the sensor scan latency of 50 ms and, concerning end-to-end latency, lower than the state of the art for this application \cite{betz2023tum}, \cite{karle2023multi}.

The algorithm can actually run at 38 Hz. As the majority of the computational time is spent in segmenting the Point Clouds, due to the parallel processing of the sensors, this method will scale well on vehicles with multiple LiDAR sensors, provided that sufficient CPU cores are available.    

\subsection{Point Cloud Segmentation}
We divided the Evaluation of the Point Cloud Segmentation algorithms into its two main steps: Ground Removal and Clustering, to understand better how potential over and under-segmentation issues can affect the vehicle tracking performance.

\subsubsection{Ground Removal} 
By proving the effectiveness of our method in removing the ground, we prove the claim regarding the high detection range and robustness of our algorithm, as effective obstacle detection requires a reliable separation of obstacles from the ground.

The Ground Removal problem can be treated as a binary classification problem, with the positive class being the non-ground objects (cars, walls, obstacles) while the negative class being ground and noise/outliers. Therefore, its performance can be evaluated by means of the classical True Positive Rate (TPR), Positive Perceived Value (PPV), and F1-score metrics.

In order to determine these metrics for every LiDAR Point Cloud in the dataset, we developed an offline automatic labeling algorithm consisting of three steps: first, the normal versor of every scan point is computed by fitting a plane to its neighbors. Then, by imposing a suitable threshold over the components of the normal versor, all the points with a vertical normal are labeled as \textit{ground}, and all the other points (belonging to the wall, opponents, or objects outside the track area) are labeled as \textit{obstacle}. The final step consists in using the K-NN labeling algorithm to remove any outliers.

With the reconstructed ground truth, it is possible to compute the TPR, PPV, and F1-score metrics for every LiDAR scan and to compare the proposed algorithm with a benchmark. As a benchmark, we chose the 2.5D binary grid approach presented in \cite{himmelsbach2009real}, representing a widely accepted solution to the Ground Removal problem which has already been applied in the IAC context by \cite{jung2023autonomous}.

\begin{figure}[b]
    \centering
    \includegraphics[width=\linewidth]{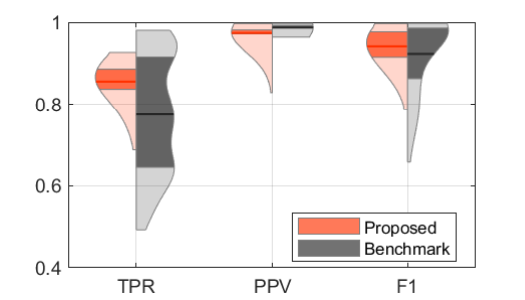}
    \caption{Violin plots of the True Positive Rate (TPR), Positive Perceived Value (PPV), and F1-score scan-wise performance of the proposed ground segmentation method against the benchmark presented in \cite{himmelsbach2009real} over the reconstructed ground truth. The distributions are cropped to the 95th percentile, with the 25th and 75th percentiles and the mean value highlighted. }
    \label{fig:res-ground-removal}
\end{figure}

Figure \ref{fig:res-ground-removal} shows the experimental results of the two algorithms over the labeled dataset. The distribution of the metrics shows how the proposed algorithm outperforms the benchmark in terms of TPR, where the benchmark shows low consistency in its performance while it underperforms in terms of PPV. The F1-score of the proposed method is larger and more consistent, showing superior overall performance. For this application, a higher PPV translates into robustness to outliers, as fewer non-ground points are labeled as ground, while TPR directly correlates with the vehicle detection performance, both in terms of detection range and number of points per vehicle. 


Since this is only the first step of the LiDAR-based perception pipeline, we considered it favorable to have a better performing algorithm as there are several additional processing steps that can remove the false positives, but there is no chance of regaining a false negative. Therefore, it is possible to conclude that the proposed method outperforms the benchmark for the metrics relevant to our application.

\subsubsection{Clustering}
\begin{figure}[t]
    \centering
    \includegraphics{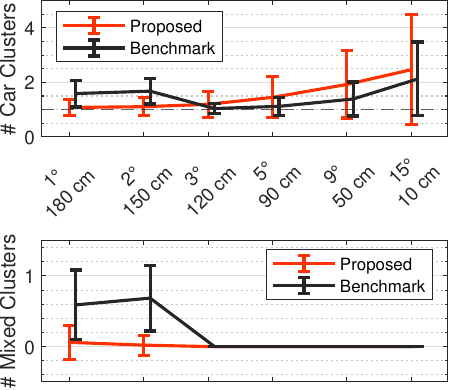}
    \caption{Clustering performance computed as average and standard deviations of the number of clusters (a) per car and mixed car-wall clusters (b) for different parameter values, respectively the $\beta$ angle for \cite{bogoslavskyi2017efficient} and the cluster size for \cite{rusu20113d}. The two curves are shifted horizontally for visualization purposes.}
    \label{fig:res-clustering}
\end{figure}


The main metric used in literature to evaluate the performance of a clustering/segmentation algorithm is the Intersection over Union (IoU) metric. However, the computation of this metric requires a segmented ground truth, which requires significant effort to build manually and proved to be very sensitive to offsets and delays when trying to build it automatically.
Furthermore, for this application, we are only interested in the correct segmentation of the points belonging to the opponent vehicles. 

Therefore, we preferred to use a pair of custom metrics to evaluate the performance of the proposed algorithm for this scenario. We then compared the proposed Range-Image approach \cite{bogoslavskyi2017efficient} with the standard Euclidean Clustering algorithm from \cite{rusu20113d}.

The first evaluation metric consists om the number of clusters containing at least one vehicle point. As only one vehicle is present in the dataset, the ideal value of this metric is one, with larger values showing a tendency of the algorithm to over-segment the car.
The second metric, on the other hand, aims to measure the tendency of the algorithm to under-segment the opponent vehicle Point Cloud as it counts the clusters which contain both points belonging to the vehicle and to other non-vehicle obstacles (usually, the wall).

As the vehicles competing in the IAC are equipped with the same sensor configuration, the timestamped RTK GNSS pose history of the two cars can be synchronized and smoothed in order to build a ground truth of the $x$ and $y$ position and differential velocity $v$ and heading $\psi$. 
These two metrics are easy to compute: by using the synchronized EGO and Opponent RTK GNSS poses, is it possible to compute the amount of clusters falling entirely inside an over-dimensioned opponent bounding box, which works under the assumption of a single opponent being present on the track, with no other obstacles. On the other hand, a mixed cluster will have dimensions much larger than a single car.

Figure \ref{fig:res-clustering} shows the results of this comparison on the labeled dataset by varying the main parameters of the two algorithms: the $th_{seg}$ angle for the Range Image Segmentation and the maximum distance for the Euclidean Clustering. It shows a clear tendency of the proposed algorithm to over-segment the car, however with considerably fewer mixed car-wall clusters.

For this application, over-segmenting the car usually results in having the Point Cloud of a wheel separated from the main vehicle body, which does not significantly impact the classification and opponent pose estimation tasks. On the other hand, a mixed car-wall cluster would be a disaster for the rest of the perception pipeline. Additionally, the dataset was acquired with a single opponent vehicle, but if we consider the scenario of two racecars engaging in a close overtaking maneuver, the danger of under-segmenting the two vehicles into a single cluster would severely impair the tracking performance.

For these reasons, we decided to empoly the Range Image clustering algorithm, with the $th_{seg}$ angle set to 2.5° in order to reduce the effects of over-segmentation as much as possible without risking having mixed car-wall clusters.

\subsection{Vehicle Detection}

\begin{figure}[b]
    \centering
    \includegraphics{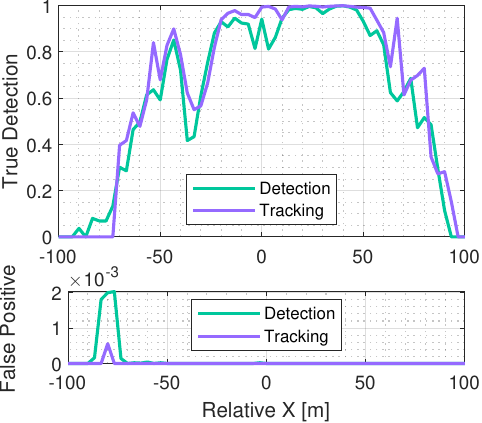}
    \caption{LiDAR detection (green) and Target Tracking confirmation (purple) probabilities given the real opponent longitudinal distance from the EGO vehicle. The top graph represents the probability of detecting an opponent at a certain distance (higher=better). The bottom graph represents the probability of a detection at a certain distance being a False Positive (lower=better).}
    \label{fig:res-detection-probability}
\end{figure}

The green curve in Figure \ref{fig:res-detection-probability} shows the probability of true and false detections as a function of the opponent (or false positive) longitudinal distance. The opponent detection probability is obtained  by dividing the number of opponent detections at a given range by the total number of LiDAR scans where the opponent is at the same distance (computed using the GNSS ground truth) whether it is detected or not. It is effectively a True Positive Rate (TPR). The false positives are normalized over the total number of LiDAR scans.

Figure \ref{fig:res-detection-probability} shows how this algorithm manages to achieve a detection range up to 90m ahead and 85m behind the EGO vehicle, with a 50\% detection probability of a vehicle located 80m ahead in the traveling direction. The effect of the blind spot in the LiDAR Point Cloud due to the rear wing of the EGO vehicle can clearly be seen from the great reduction in the opponent detection probability in the [-20m,-40m] range.

The false positive detections are mostly located in the [-70m,-90m] range, as they are mainly caused to the EGO heading estimation error. Due to the effect of the car sideslip, the estimated vehicle heading coming from the differential GNSS position history tends to be closer to the vehicle's course direction than to the effective heading of the vehicle body. This leads to objects outside the racetrack limits being projected inside the track area when consulting the map.

\subsection{Vehicle Pose Estimation and Tracking}

\begin{figure*}
    \centering
    \includegraphics[width=\textwidth]{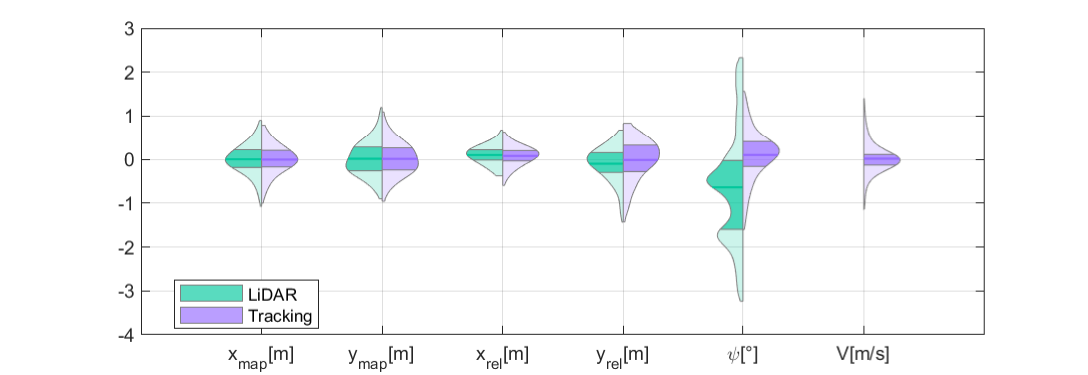}
    \caption{Violin plots of the raw measurements and tracked state estimation error computed with respect to the ground truth. The $x$ and $y$ position errors are computed both on the map and the vehicle-fixed reference frames. The distributions are cropped to the 95th percentile, with the 25th and 75th percentiles and the mean value highlighted.}
    \label{fig:res-error-violin}
\end{figure*}

Figure \ref{fig:res-error-violin} shows the distribution of the LiDAR measures and the Tracking state estimation error computed with respect to the GNSS ground truth. For this analysis, we used only the measurements and tracks corresponding to the opponent, obtained by segmenting the dataset using the Cartesian distance from the ground truth.

The analysis of the errors in the local reference frame is particularly helpful in identifying the bias present in the relative $x$ direction, which can be due to sensor miscalibration. On the other hand, the non-zero mean, high variance, and asymmetrical distribution in the relative $y$ error show an over-representation of the negative values. This can be explained once again by the effect of the significant vehicle side slip on the EGO state estimation error. Since the dataset is composed only of left turns, a positive sideslip would result in an artificial bias of the measurements towards the right of the EGO vehicle (negative $y$). The $\psi$ error shows a clear bivariate distribution, coming from the system switching between the outputs of the double estimators.

The effect of the tracking filters helps to reduce the bias in the local reference frame. The chosen tuning of the EKF covariance matrices aims to have a smooth estimation of the opponent velocity to allow for precise planning of overtaking and avoidance maneuvers by sacrificing the precision on the $x$ and $y$ position while maintaining the estimation error for all the states under acceptable levels.

Another beneficial aspect of the target tracking algorithm is the added persistence and outlier detection capabilities obtained by the track management logic. The purple Curve in Figure \ref{fig:res-detection-probability} shows how the track management logic allows to continue estimating the opponent state even in the regions subject to a low detection probability due to the LiDAR blind spots.
At the same time, the probability of tracking a false positive detection is greatly reduced.

\section{Conclusion}

This work presented an online LiDAR-based vehicle detection and tracking algorithm for autonomous racing, composed of multiple modules solving many crucial problems of vehicle detection from Point Cloud data. The algorithm underwent extensive field testing during the 2023 IAC @ CES competition, which involved head-to-head overtaking maneuvers at the Las Vegas Motor Speedway. Notably, this algorithm allowed team PoliMOVE to successfully overtake a target vehicle moving at 250 km/h.

Regarding Point Cloud Segmentation, the proposed approach outperformed a benchmark method in terms of TPR and F1-score. The ground removal algorithm effectively distinguished non-ground objects from ground and noise/outliers, leading to robust opponent vehicle detection and tracking. The clustering algorithm exhibited a tendency to over-segment opponent vehicles, which is acceptable considering the ability to filter false positives in later stages of the pipeline.

The precise clustering together with the map-based approach helps the algorithm to ensure a considerable detection range while avoiding the loss of track of the opponent vehicle. Although this came at the cost of more false positives, they were mainly due to objects outside the track surface and could be effectively filtered.

Measure Extraction and Target Tracking further improved the accuracy of the opponent vehicle's state estimation. The tracking algorithm demonstrated good performance in filtering and predicting the opponent's position, velocity, and heading. The use of an Extended Kalman Filter with appropriately tuned covariance matrices provided smooth and precise opponent velocity estimation, crucial for accurate planning of overtaking and avoidance maneuvers.

Despite challenges such as LiDAR blind spots and EGO state estimation errors affecting the performance, the proposed algorithm exhibited low tracking error and high outlier robustness. It is capable of tracking an opponent up to 80 meters ahead, with an average processing latency of 26ms.

In conclusion, our method demonstrates superior performance over current algorithmic approaches in vehicle detection and tracking. Future advancements in this domain are anticipated to be centered around Neural Network-based techniques, particularly for vehicle detection and pose estimation using raw or segmented LiDAR data. Importantly, the modular design of our algorithm allows for seamless integration of CNN components while preserving the overall architecture.\label{sec: conclusions}





\section*{Acknowledgment}

The authors would like to thank all the past and present members of the PoliMOVE Autonomous Racing Team, especially Marco Mandelli for his inputs in the conception and setup of this research, Andrea Marcer for his help in the software architecture design and online implementation, and Brandon Dixon and Robert Cole Frederick for their indispensable assistance in engineering and operating an autonomous racecar.

Furthermore, a special thanks goes to the TUM Autonomous Motorsport team for their role in the acquisition of the data used in this work and for motivating us to constantly improve our research, race after race, and to all the Indy Autonomous Challenge staff and teams for providing us with this unique research opportunity.


\begin{table}[h!]\label{tab: params}
  \caption{List of parameters and respective values used in the experimental performance evaluation}

  \centering
  \begin{tabular}{| c | c | l |}
    \hline
    \textbf{Symbol}         & \textbf{Value}  & \textbf{Description} \\
    \hline
    $n$                     & 32              & Range/Elevation Image Rows \\
    $m$                     & 857             & Range/Elevation Image Columns \\
    $sz_{SG}$               & 5               & Sawitsky-Golay filter kernel size \\
    $th_{gnd}$              & 20.0°           & Threshold for Ground Segmentation \\
    $th_{seg}$              & 2.5°            & Threshold for Range Image Clustering \\
    $w_s{sz}$               & 9               & Windows Size for NGRIR \\
    $th_r$                  & 5.0m            & Threshold on Range for NGRIR \\
    $th_{mrg}$              & 1.80m           & Threshold for Cluster Merging \\
    \multirow{2}{*}{$Q_{diag}$} 
                            & $\{0.005, 0.005, 0.5, $ 
                                              & \multirow{2}{*}{EKF state error covariance} \\
                            & $0.005,0.0005\}$  & \\
    $R_{diag}$        & $\{5.0, 5.0\}$ & EKF measurement error covariance \\
    $N$                     & 20              & Iterations in track management \\
    $M_c$                   & 6               & Confirmation Threshold \\
    $M_e$                   & 5               & Elimination Threshold \\

    \hline
  \end{tabular}
  
\end{table}

\ifCLASSOPTIONcaptionsoff
  \newpage
\fi



\bibliographystyle{IEEEtran}
\bibliography{bibliography/IEEEabrv,bibliography/IAClidarTracking}









\end{document}